# Dzongkha Next Word Prediction System

Prerna Chhetri*[1], Tenzin Yoezer[2], Phuntsho Wangmo[3], Tshewang Bomjan[4].

Information Technology Department, College of Science and Technology, Royal University of Bhutan,
Email: 02200180.cst@rub.edu.bt*[1] , 02190165.cst@rub.edu.bt[2],
02200160.cst@rub.edu.bt[3] , 02200170.cst@rub.edu.bt[4]

## Abstract

Dzongkha, being the national language of Bhutan is a common and widely spoken language in the country. Official documents, scriptures and other literature products are written in Dzongkha Language in order to retain the cultural value. However, documenting Dzongkha writing is a challenging and time-consuming process, largely due to the complexity of the script, the need for multiple keystrokes per syllable, and the limited availability of efficient typing tools. An immediate system that can predict and display a list of probable words for Dzongkha Language is the solution for this problem. The project is mainly aimed to make Dzongkha typing as convenient as possible by reducing the number of keystrokes. The system can prove to be beneficial especially for Dzongkha writers. Our dataset is acquired from DCDD and has a total of 100000 sentences, 1331282 words and 28344 unique words. The data preprocessing was done by removing all the alphanumeric characters, tokenization, generating N-gram sequence and padding. Three models that were selected for training are LSTM, Bi-LSTM and GRU. The training process included fine-tuning of the model's hyperparameters. Accuracy is the performance metrics which we used to determine the performance of each model. While training the model on a 100000 sentences dataset, GRU being lightweight and its ability to handle larger dataset performed better with 74.03% accuracy and also solved the problem of overfitting.

## 1. INTRODUCTION

Dzongkha is the national language of Bhutan. People belonging to different regions of Bhutan have their own dialect which was spoken from their ancestral times. These people use Dzongkha as a language to communicate with people of different regions and dialects as it is a common language in the country. Moreover, it is also a standard language that is spoken in the various government institutions and offices.

Dzongkha language typing is very time consuming and also prone to a lot of errors. To address this, next word prediction systems offer a potential remedy. By suggesting Dzongkha words and reducing manual effort, these systems enhance typing speed and accuracy. This aids in preserving language authenticity. Implementing such technology can alleviate typing challenges, maintain language richness, and promote meaningful expression in Dzongkha.

Even though Dzongkha is being promoted through writings, messages and documentations, it is evident that Dzongkha typing is difficult (Wangchuk et Al., 2021). Unlike typing alphabetic scripts, in Dzongkha each syllable is composed of multiple characters and because of that one would require to use more key presses to type the words which can be quite hectic and time consuming. Not to mention Dzongkha spelling is unnecessarily cumbersome.

Keeping all these issues in mind, there is an urgent need for a system for Dzongkha writers that can make Dzongkha typing as convenient as possible. For these reasons, this project proposes a Dzongkha Next Word Prediction System that is able to predict the next syllable, reduce keystrokes and enhance the experience of Dzongkha typing.

## 2. LITERATURE REVIEW

The paper written by Wangchuk et al., mentioned that their next syllable prediction system aimed to address all the issues concerned with Dzongkha typing, including the use of several keystrokes. An LSTM neural network was incorporated to make the model learn. The model learns the syntactic and semantic features of language through the sequence generated from the Dzongkha sentences. The model had an accuracy of 78.33 percent, and deployment was done using the Django system (Wangchuk et al., 2021).

A Dzongkha next syllable prediction system is mainly aimed to reduce keystrokes and key presses in order to make Dzongkha typing fast and as convenient as possible. Segmented datasets were provided by the Dzongkha Development Commission of Bhutan, and these were processed through variants of RNN models. Bi-LSTM model proved to be the most suitable model with two hidden layers and 512 hidden layer neurons, which contributed to an accuracy of 73.89 percent (Wangchuk et al., 2023).

In another study most evident architectures and regularization methods with extensive automatic black box hyperparameters ruining were evaluated. After the evaluation it was found that standard LSTM architectures had gone through proper regularization and it had the best performance from the nominated models (Melis et al., 2018).

A study conducted by Gannai & Khursheed, says that Recurrent Neural Networks (RNNs) represent the forefront of processing consecutive data, being used in technologies like *Apple's Siri* and *Google Voice Search* and LSTM. The feedback connections in it, acts as a versatile tool, which is able to work with single data points like images and entire sequences such as speech or video. This flexibility is clearly evident in tasks like recognition of connected handwriting and speech understanding. A fundamental aspect of NLP is that the N-gram Language Model predicts word groups one after another, while neural networks play a crucial role in predicting next words in a language model. (Ganai & Khursheed, 2019)

The next word prediction for Hindi was done using two deep learning techniques: LSTM and Bi-LSTM (Sharma et al., 2019). These neural networks predict the next word of a given sequence of at least six words. Both models were implemented using the categorical cross-entropy loss and softmax activation functions. Both of these models were further trained with a learning rate of 0.001. The accuracy of the LSTM was 70.89%, and the accuracy of the Bi-LSTM was 79.54%. The validation accuracy for the two models is, respectively, 59.46% and 81.07%. For the same dataset, Bi-LSTM model performs better and learns more quickly than LSTM.

# 3. METHODOLOGY

## 3.1. Project Methodology

For Literature Review, various authentic and relevant papers will be reviewed and the concepts learnt will be applied throughout the span of the project. Data Acquisition is done to collect a corpus of text in Dzongkha, which will be acquired from the DDC (Dzongkha Development Commission) Data Preprocessing's Specific techniques such as tokenization and normalization is done to improve the precision and performance of models. Word embedding is a technique where words are represented as vectors of real numbers. These vectors are typically learned from a corpus of text, and they can be used to represent the meaning of words,their relationships to other words, and their context in a sentence. Some of the word embedding techniques that are usually implemented are Bag-of-words, TF-IDF, Word2Vec and Glove. For Model Training, two or more models will be selected and trained on the same dataset. Then the model that gives us the highest accuracy or optimal result will be selected. If the model doesn't give us the desired output, then fine tuning of the hyperparameters will be done. The results of the fine-tuned parameters will then be sent for word embedding. The process will continue until the desired output is obtained. For Framework selection, an appropriate framework will be selected for the development of the system. The selected framework will aid in the easy integration of the model with the system in the later phase. For GUI Design, a user-friendly interface(prototype) will be designed for the system which will be used as a reference during the front-end development stage. In the Development stage, the system components are coded/implemented according to the design of the GUI using the selected framework.

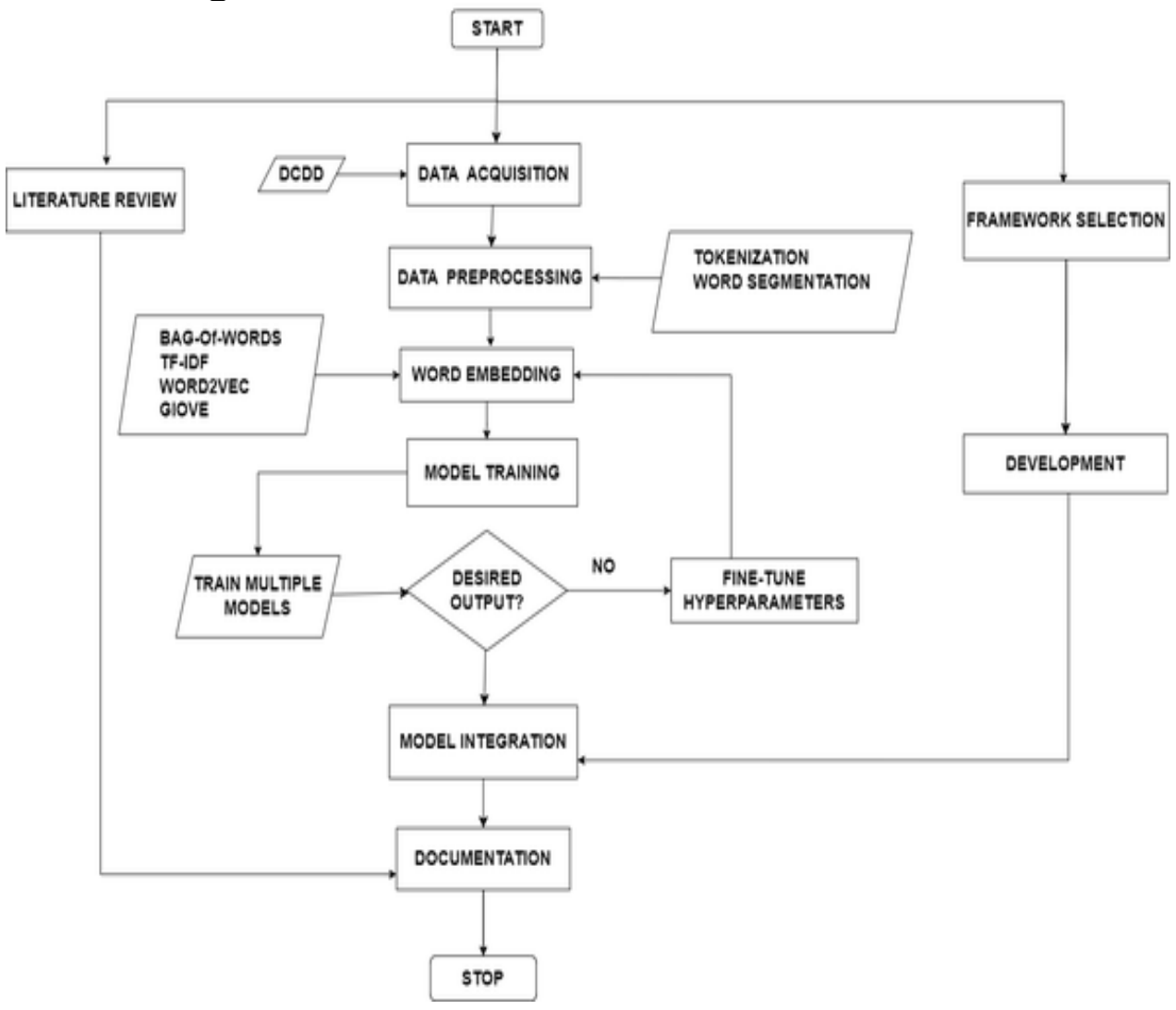

Fig 1: Project Methodology flowchart

In the Development stage, the system components are coded/implemented according to the design of the GUI using the selected framework. In the Model Integration phase, the development of the system and model training completed, the model will be installed in the system. In the Documentation phase the user manuals, technical documentation and training materials are created. In Deployment phase, the system is deployed and made available to the users.

### 3.2 System Requirements

Outlined below are the system requirements for the implementation of Dzongkha next word Prediction system, encompassing both hardware and software aspects. The detailed requirements for undertaking such a project are divided into two categories. The hardware requirements are minimum of 8GB ram and the most preferred is 16GB or higher, Intel Core i5 or i7,500 GB storage, GPU NVidia RTX 3090 (48 GB VRAM).The software requirements are Deep Learning library- Keras, Tensorflow, Pytorch, Huggingface, Python3, JavaScript, HTML, CSS, GoogleColab, Figma, Django Framework, Microsoft excel, csv.

### 3.3 Technology used

We used figma to design a prototype for our web app, .HTML, CSS and javascript for the frontend development. BOOTSTRAP was used to make our web app responsive. Django, a python-based web framework was used for the backend development and SQLite is used as the database. Model training was done using google colab.

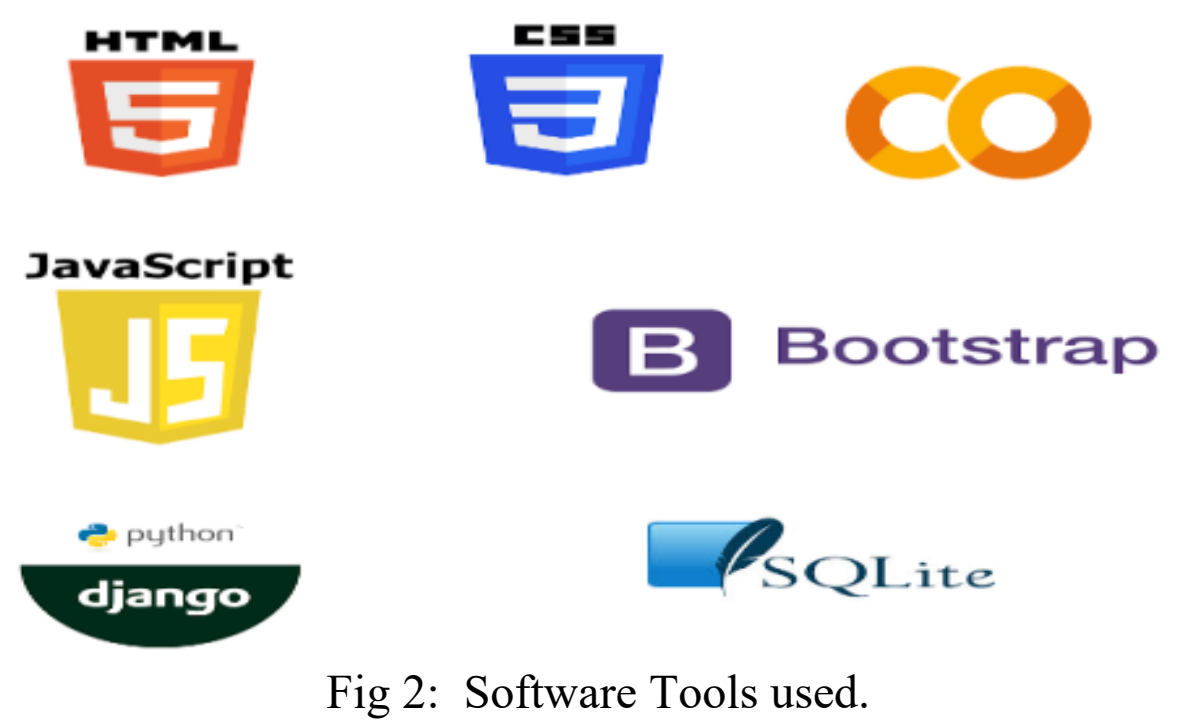

Fig 2: Software Tools used.

## 4. RELATED WORKS

### 4.1 Dataset Preparation

At the time of training the model to start with there were many constraints regarding the size of data we could take because we were not having a computer system which could hold a large data set.

We then increased the size of the dataset and utilized the GPU system for training our model. The data preprocessing was done by removal of alphanumeric characters, '#' ,'?','།', (་) at the word level. We remove alphanumeric characters, special characters, '།',?. Tsha and shay are removed because we want to make the Dzongkha words equivalent to English words. We remove alphanumeric characters, special characters because in a word, it doesn't contain special characters and symbols. Non-alphabetic and numeric characters may not provide meaningful information for the language model and can introduce unnecessary complexity.

Table 1: 20,000 Dataset

| Sl No | Variable | Value |
|---|---|---|
| 1 | Total Sentence | 20000 |
| 2 | Total words | 559,344 |
| 3 | Total No of Unique Words | 7816 |
| 4 | No.of sequence | 115,351 |

Table 2: 100,000 Dataset

| Sl No | Variable | Value |
|---|---|---|
| 1 | Total Sentence | 100000 |
| 2 | Total words | 1,331,282 |
| 3 | Total No of Unique Words | 28,344 |
| 4 | No.of sequence | 623,820 |

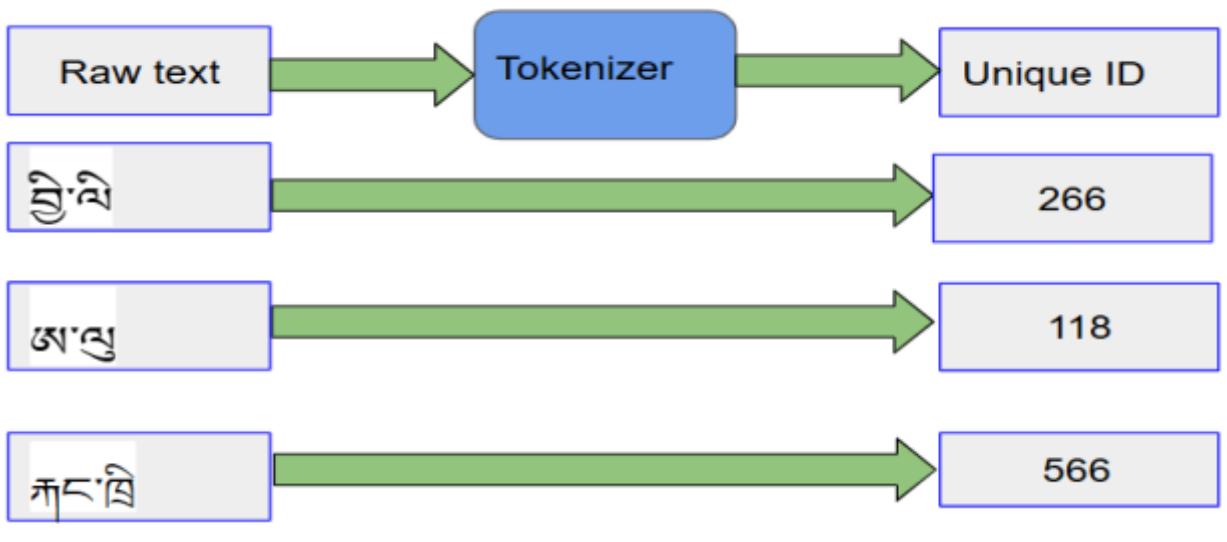

Fig 3: Tokenization Process

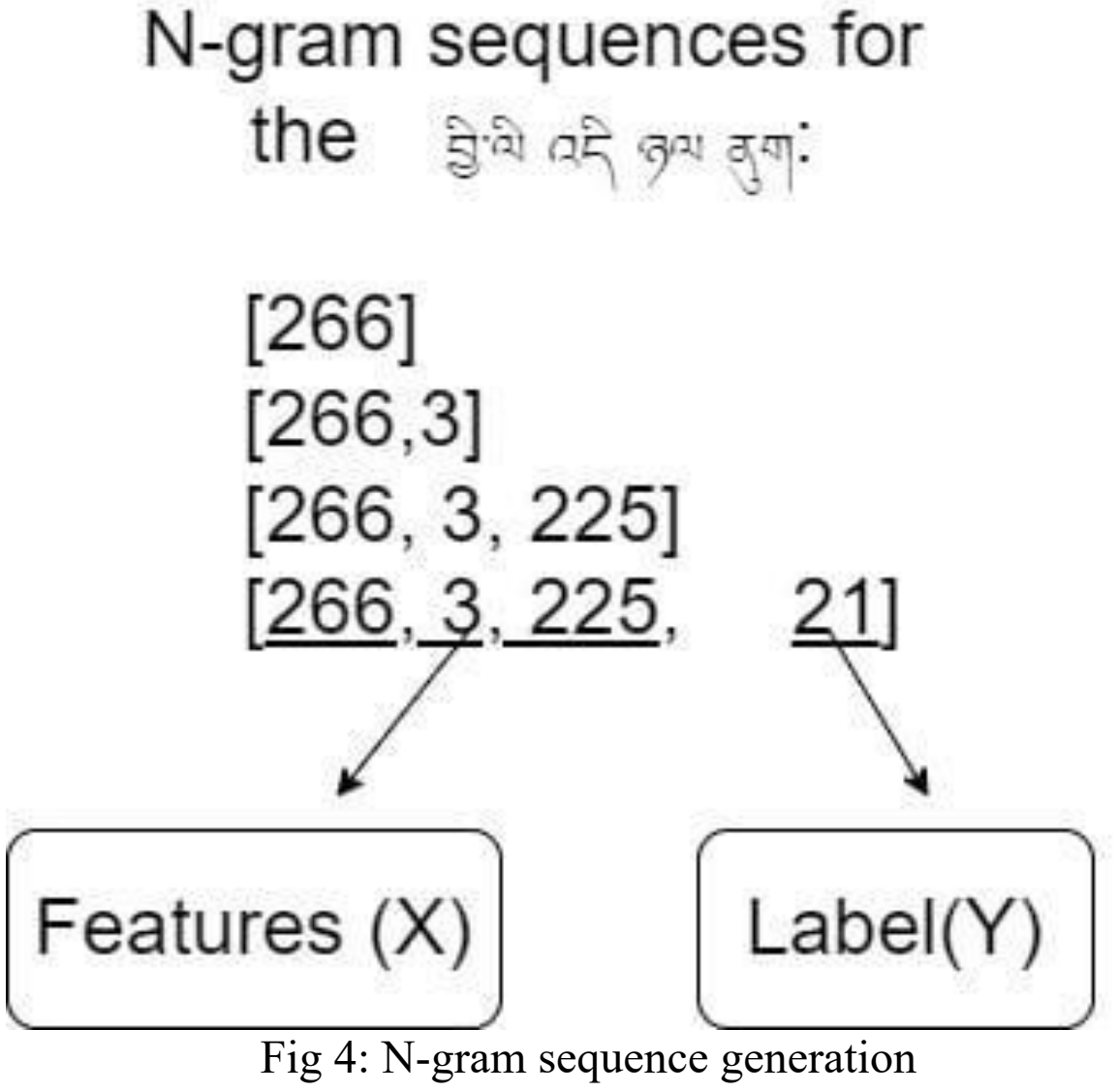

Fig 4: N-gram sequence generation

Tokenization is a pre-processing task that has the goal of segmenting the text into smaller unit. We employ a successive tokenization method that comes from TensorFlow to divide every of the Dzongkha word depending on the vocabulary list of every unique word accompanied with its unique integer values. The figure 4 depicts the raw text that is tokenized with a unique ID. For example, we have ‘བྲི་ལི’ which is tokenized with a unique ID of 266.

We used the N-gram sequence for sequence generation. The N-gram sequence is created in order to curate the dataset into features(X) and labels(Y). As we can see the last token in each sequence is Y(label) and all the other tokens before Y are X or the features. The model is trained to predict Y based on X. where y is label and X is the features.

In the DzoSyll dataset, the string lengths range between 1 and 171 letters or characters for each term in the dataset. We decided not to expand the maximum sequence length to the length of longer sentences, while the majority of the sentences in our dataset train and dev set had a maximum sequence length of 22. To be able to construct matrices with equal dimensions for all input sequences, the n-gram sequences are completed by zeros in the front. Padding is carried out to make sure that the feature sets of all the inputs are equal in size and neural networks demand inputs of equal dimensionality. As a result of tokenization, it is necessary to append additional tokens for each sequence to reach the length of the longest sequence in the corpus. Adding a beginning special token also allows the model to ignore all the other unnecessary data, also preserves the time order and help the model to understand the relationships between words and context.

The maximum sequence length is 22 so in order to make the length of the input sequences equal to the maximum length, pre-padding needs to be done. Pre-padding the process of adding zero bits at the beginning of the sequence so as to make the length equal to the maximum length.

### 4.2 Model Training

Accuracy is chosen primarily since test metrics that maximize accuracy were chosen to evaluate the models in terms of how accurate they make the predictions. Some of the hyperparameters that we chose are Epoch, Batch size and validation split. Epochs is the utterance of the feed forward and the number of cycles the entire data set has to pass through the neural network. Batch size defines how many training samples the network should be passed across before updating the model. Validation split is done in order to hold out some of the data for the validation set. In training, the model will not make an attempt to train on this data but rather assess its performance on it.

The training of the model is crucial to achieve a well-performing language model. The function create_model defines the architecture of a deep learning model for text prediction using the TensorFlow Keras framework. Sequential Model forms a linear stack of layers, the hallmark of the crust. We chose the Sequential class because it is one of the easiest ways to create a single neural network, where the model has one input and one output, and hence, every layer has one input tensor and one output tensor. Embedding Layer is a layer in natural language processing models, this encompasses an input layer, a hidden layer, and an output layer. It transforms significant, diagonally stored vectors of integer representations of words with many zeros to small and dense vectors. First GRU Layer is a layer is a recurrent one with 512 units to deal with sequences through gated recurrent units (GRUs).It should be noted that GRUs are a type of RNN and are most commonly used for sequence prediction tasks. The Second GRU Layer turns out to be another GRU layer, however, on this occasion with 256 units. This time, the setting of return_sequences=False will be used to have the layer spit out only the final output in the output sequence, which in effect turns the sequence into a vector. In this case, in the dense layer, the number of neurons in this layer is 100 plus one fourth of the size of the vocabulary list, and this is a fully connected layer. Output Layer consists of a single layer of dense neurons equal to the vocabulary's number of words. A softmax transformation will retrieve the probability distribution of the next word from the word dictionary. The output of each neuron in the network is the probability of the transition to the next word, which corresponds to the particular neuron. For Compilation, the model is trained using the loss function, the definition of which is ideally referred to as 'sparse categorical cross-entropy loss' for problems where each instance is expected to belong to one class exclusively.

Fine-tuning means adjusting the model after training to make it more efficient or better suited to the specifics of the data. We make changes in the number of layers in GRU or layer size by altering the number of units. This had to be done to change the learning rate of the Adam optimizer as the training progressed. We may also further experiment with different architectures or add further regularizations such as Dropout.

### 4.3 Model Integration

The integration of the machine learning model into the web application involves several key steps: The trained model is loaded into the application. The tokenizer used during the training phase is also loaded to ensure consistency in text processing. Backend functions are embedded and used to implement the processing of the user's input, prediction implementation, and the transmission of results to the frontend. The frontend is developed to communicate with the backend, capture user data, present the user with predictions, and handle user interactions.

### 4.4 Challenges and Solutions

During the integration process, several challenges were encountered and addressed. Large data were used in the model making and more than one prediction had to be made in a shortest possible time, through optimization. For ensuring accuracy, Other procedures that were useful to get various solutions specified included the fine tuning of the given model and the data preprocessing. User Interface Design led to much time being spent during the planning phase and consultations with users on how to arrive at the ergonomic and highly sensitive model.

## 5. RESULT AND DISCUSSION

### 5.1 Model Comparison for 100K dataset.

The size of the dataset was increased. The figure below shows the model performance while training it on a 100K dataset. The hyperparameters such as Epoch (200), Dense Layers (2), Model Layers(2), neuron (1024,512), Batch Size (500), Early Stopping(10) and Dropout (0.2) were used in training of the three models.

Table 3: 100k Dataset Details

| Sl No | Variable | Value |
|---|---|---|
| 1 | Total Sentence | 100000 |
| 2 | Total words | 1,331,282 |
| 3 | Total No of Unique Words | 28,344 |
| 4 | No.of sequence | 623,820 |

Table 4: Model comparison on 100k dataset

| Sl No | Model Name | Time(hours) | Accuracy(%) |
|---|---|---|---|
| 1 | LSTM | 8 | 60 |
| 2 | Bi-LSTM | 13 | 70.79 |
| 3 | GRU | 15.33 | 74.03 |

We achieved the highest accuracy of **74.03%** with the GRU model.

## 5.2 Graphical Representation of Model performance for 100K dataset.

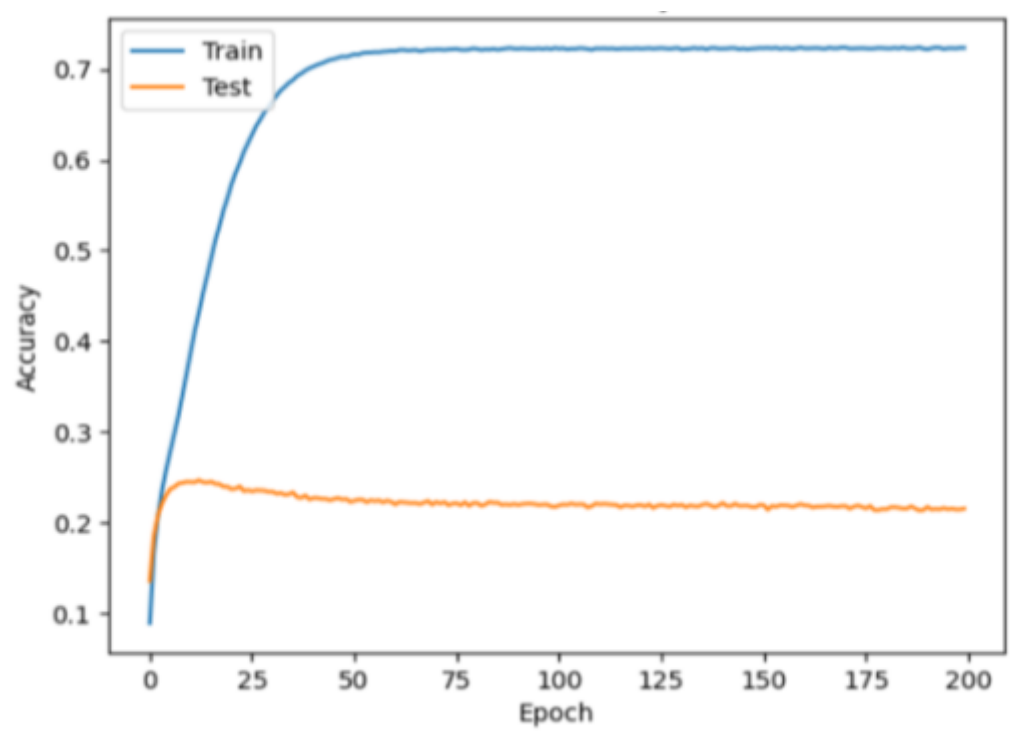

Fig 5: LSTM accuracy (100K) result in graphical form

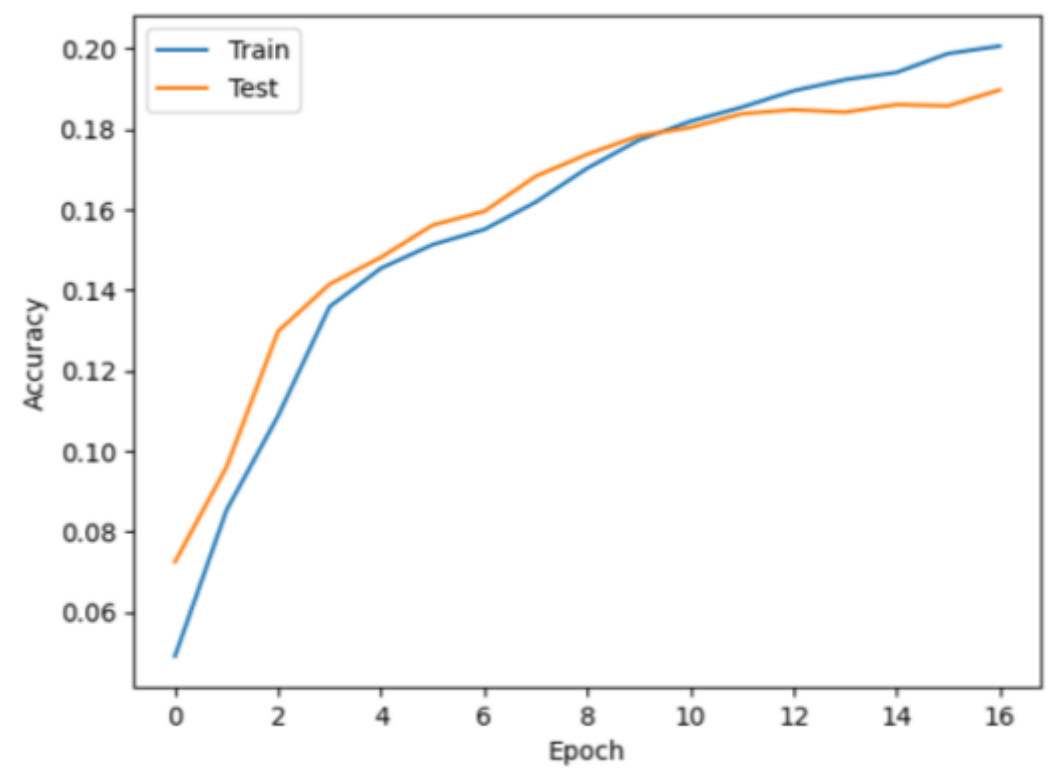

Fig 6: GRU accuracy(100K) result in graphical form

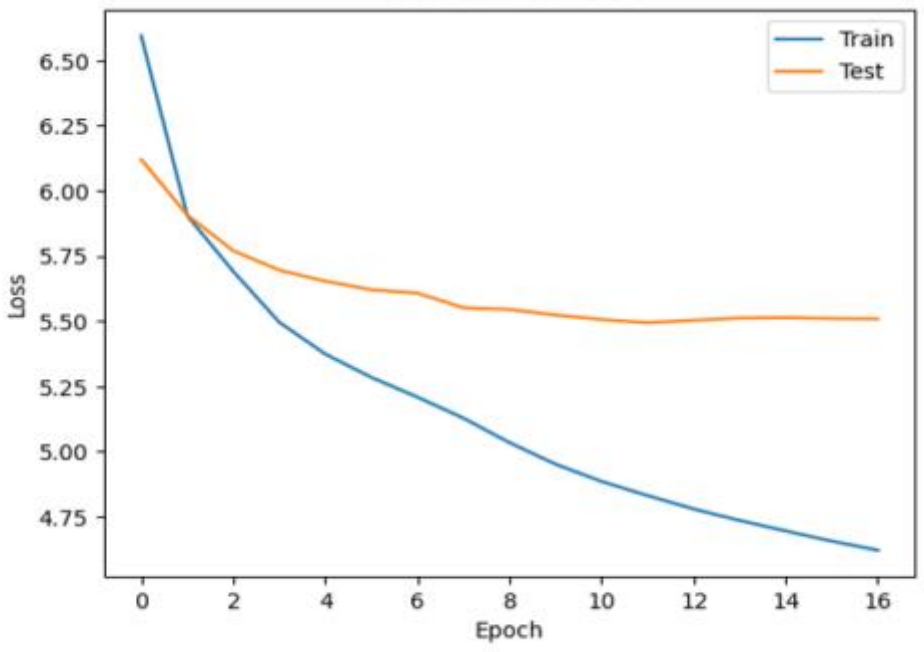

Fig 7: GRU Model loss (100K) result in graphical form

Acceptable level of accuracy was obtained while training the model on the dataset listed above but it totally solved the problem of overfitting, the model was able to make accurate predictions for the samples that were not only used for training but also for the new data samples. GRU model is used for integration with the Dzongkha Next Word Prediction Application.

## 6. CONCLUSION

In a nutshell, our project successfully implemented the GRU model architecture for Next word prediction and achieved promising results. Initially, due to lack of powerful systems for model training, we trained the model on a smaller dataset, and all the models gave us an acceptable level of accuracy such as 73.79% for LSTM, 71.14 for Bi-LSTM and 70.04 for GRU but we faced issues of overfitting. During the second time, we were equipped with GPU system for training our model, we introduced some new hyperparameters and the model hyperparameters underwent some fine tuning and the percentage of accuracy obtained were 60% for LSTM, 70.79 % for Bi-LSTM and 74.03% for GRU and it totally solved the problem of overfitting. Thus, GRU was used for integration with our frontend system. Some of our future work includes development of software package(.exe) for our web app and further improvement of the model inorder to bring about better overall performance of our app.